%% file: main.tex
\documentclass[10pt,twocolumn,letterpaper]{article}

\usepackage{cvpr}              
\usepackage{cuted}
\usepackage{url}
\usepackage[T1]{fontenc}    
\usepackage{url}            
\usepackage{booktabs}  
\usepackage{wrapfig}
\usepackage{subcaption}
\usepackage{amsmath}
\usepackage{algorithm}

\usepackage{algpseudocode}
\usepackage{graphicx}
\usepackage[table]{xcolor}  
\usepackage{tabularx}
\usepackage{graphicx}
\usepackage{amsfonts}       
\usepackage{nicefrac}       
\usepackage{microtype}      
\usepackage{xcolor}         
\newcommand{\minisection}[1]{\vspace{0.02in}\noindent{\bf #1}}
\usepackage{multirow}
\usepackage{pifont}
\newcommand{\cmark}{\ding{51}}%
\newcommand{\xmark}{\ding{55}}%
\usepackage{float}

\usepackage{arydshln}
\usepackage{enumitem}
\usepackage{makecell}
\input{preamble}
\definecolor{cvprblue}{rgb}{0.21,0.49,0.74}
\usepackage[pagebackref,breaklinks,colorlinks,allcolors=cvprblue]{hyperref}

\def\paperID{13143} 
\def\confName{CVPR}
\def\confYear{2026}

\title{WaDi: Weight Direction-aware Distillation for One-step Image Synthesis}

\author{Lei Wang\textsuperscript{1},\, Yang Cheng\textsuperscript{1},\,  Senmao Li\textsuperscript{1},\, Ge Wu\textsuperscript{1},\, Yaxing Wang\textsuperscript{1,3}$^\dagger$,\, Jian Yang\textsuperscript{1,2}$^\dagger$\\
\textsuperscript{1}{PCA Lab, VCIP, College of Computer Science, Nankai University} \\ \textsuperscript{2} {PCA Lab, School of Intelligence Science and Technology, Nanjing University} \\ \textsuperscript{3} {NKIARI, Shenzhen Futian} \\
\texttt{\small \{scitop1998,\ cyrene0613,\ senmaonk,\ gewu.nku\}@gmail.com},\,
\texttt{\small \{yaxing,csjyang\}@nankai.edu.cn}\\
Code: \url{https://github.com/gudaochangsheng/WaDi}
}

\begin{document}
\makeatletter
\g@addto@macro\@maketitle{
  \begin{figure}[H]
  \setlength{\linewidth}{\textwidth}
  \setlength{\hsize}{\textwidth}
  \centering
  \includegraphics[trim=0 0.2cm 0 2cm, width=0.65\textwidth]{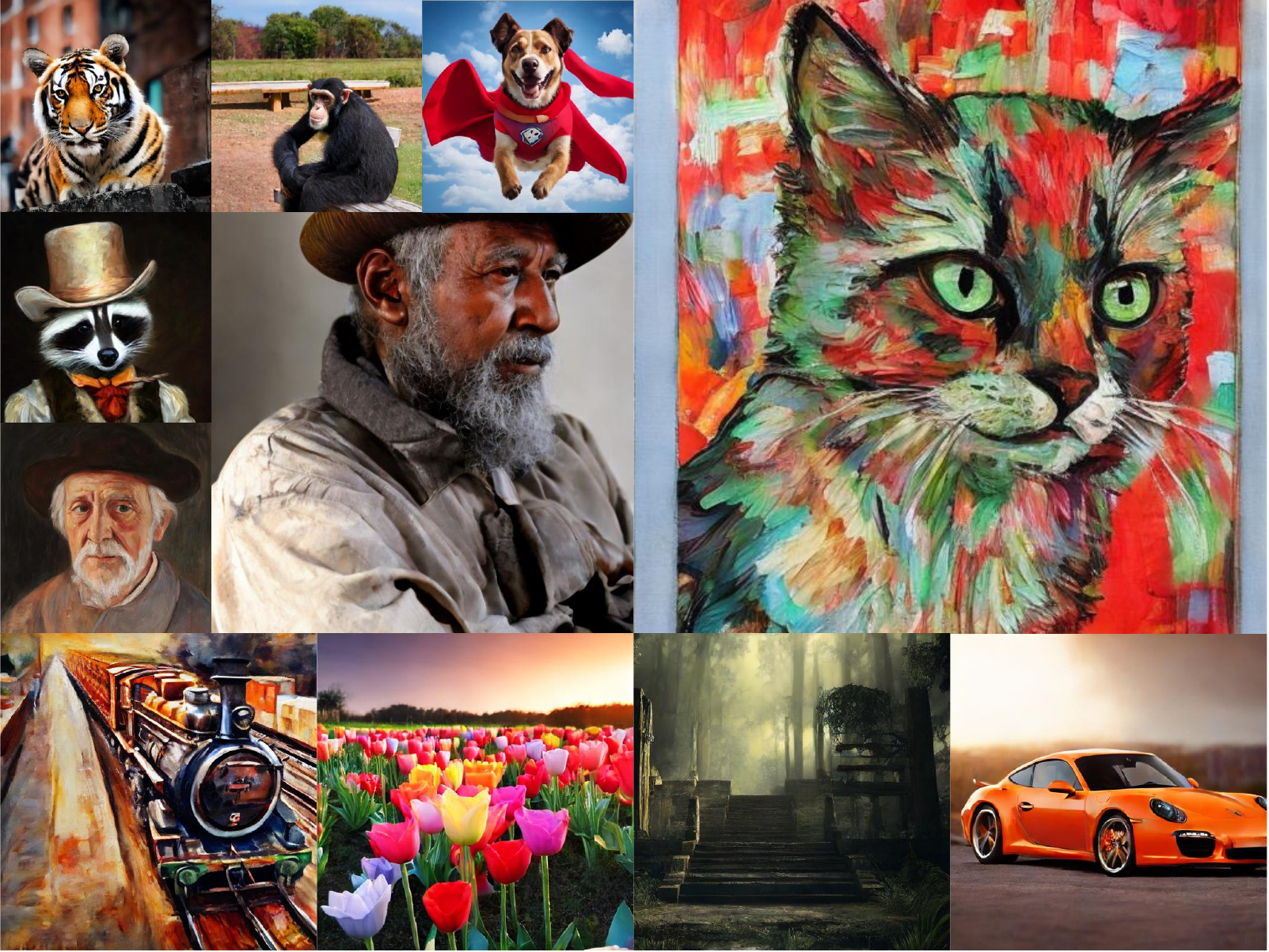}
  \vspace{-2.5mm}
  \caption{One-step generated images using our proposed method WaDi (\textit{i.e.}, SD 2.1).} \label{fig:abstract-fig}
  \end{figure}
}
\makeatother
\maketitle
\renewcommand{\thefootnote}{$\dagger$} 
\footnotetext{Corresponding authors.}

\input{sec/0_abstract}    
\input{sec/1_intro}
\input{sec/2_related_work}
\input{sec/3_method}
\input{sec/4_experiment}
\input{sec/5_conclusion}

\clearpage
\section*{Acknowledgement.}
This work was supported by the National Science Fund of China under Grant Nos, 62361166670 and U24A20330, the ``Science and Technology Yongjiang 2035'' key technology breakthrough plan project (2024Z120), the Shenzhen Science and Technology Program (JCYJ20240813114237048), the Chinese government-guided local science and technology development fund projects (scientific and technological achievement transfer and transformation projects) (254Z0102G), and the Supercomputing Center of Nankai University (NKSC).

{
    \small
    \bibliographystyle{ieeenat_fullname}
    \bibliography{main}
}


\end{document}

%% file: sec/0_abstract.tex
\begin{abstract}
Despite the impressive performance of diffusion models such as Stable Diffusion (SD) in image generation, their slow inference limits practical deployment. Recent works accelerate inference by distilling multi-step diffusion into one-step generators. To better understand the distillation mechanism, we analyze U-Net/DiT weight changes between one-step students and their multi-step teacher counterparts. Our analysis reveals that changes in weight direction significantly exceed those in weight norm, highlighting it as the key factor during distillation. Motivated by this insight, we propose the \textbf{L}\textbf{o}w-rank \textbf{R}ot\textbf{a}tion of weight \textbf{D}irection (LoRaD), a parameter-efficient adapter tailored to one-step diffusion distillation. LoRaD is designed to model these structured directional changes using learnable low-rank rotation matrices. We further integrate LoRaD into Variational Score Distillation (VSD), resulting in \textbf{W}eight Direction-\textbf{a}ware \textbf{Di}stillation (WaDi)—a novel one-step distillation framework. WaDi achieves state-of-the-art FID scores on COCO 2014 and COCO 2017 while using only approximately 10\% of the trainable parameters of the U-Net/DiT. Furthermore, the distilled one-step model demonstrates strong versatility and scalability, generalizing well to various downstream tasks such as controllable generation, relation inversion, and high-resolution synthesis.
\end{abstract}

%% file: sec/1_intro.tex
\begin{figure*}[htbp]
  \centering
  \begin{subfigure}[c]{0.52\linewidth}
    \centering
    \includegraphics[width=\linewidth]{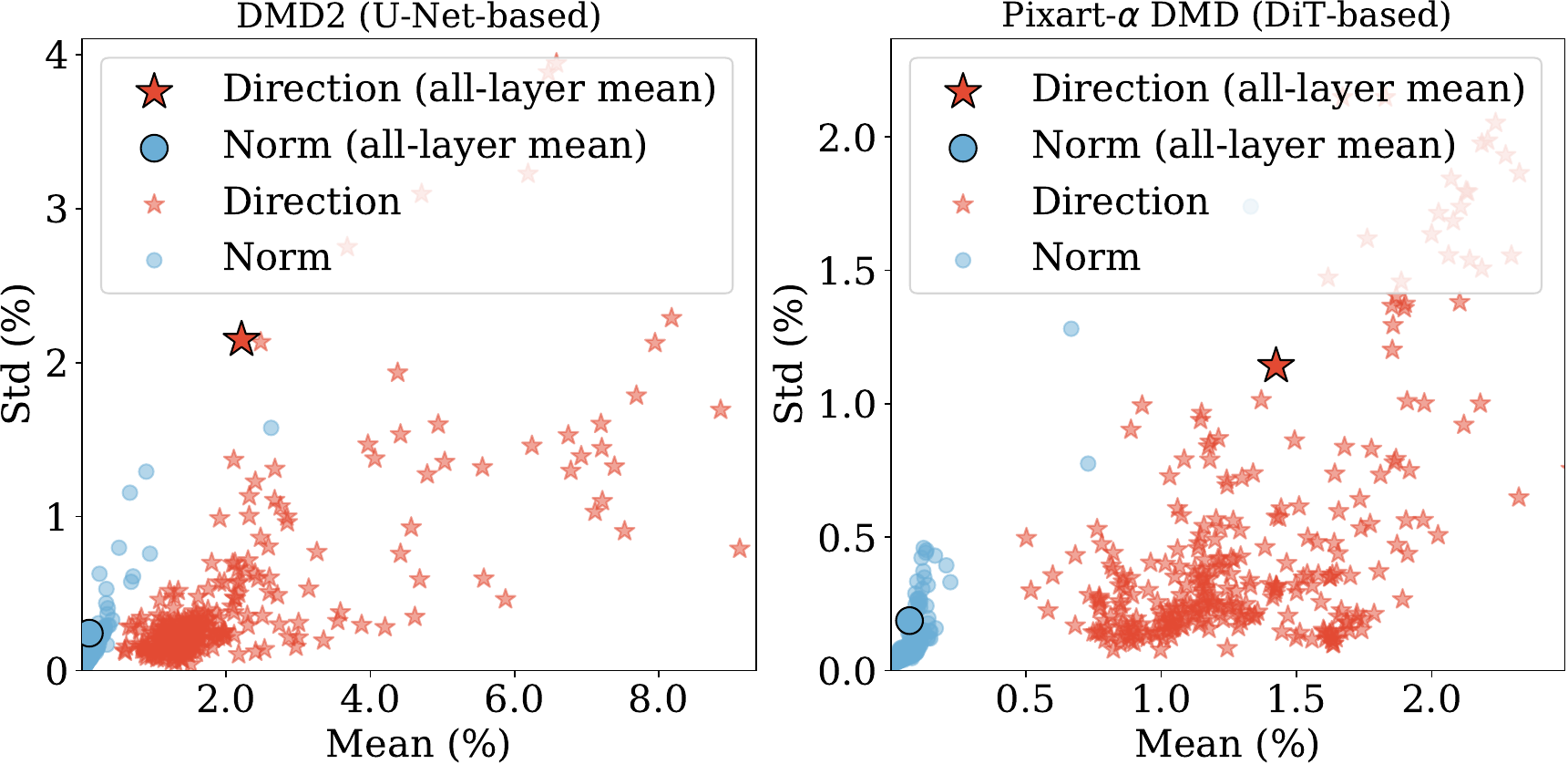}
    \vspace{-0.55cm}
    \caption{}
    \label{fig:mot-1}
  \end{subfigure}
  \hspace{0.03\linewidth}
  \begin{subfigure}[c]{0.35\linewidth}
    \centering
    \includegraphics[width=\linewidth]{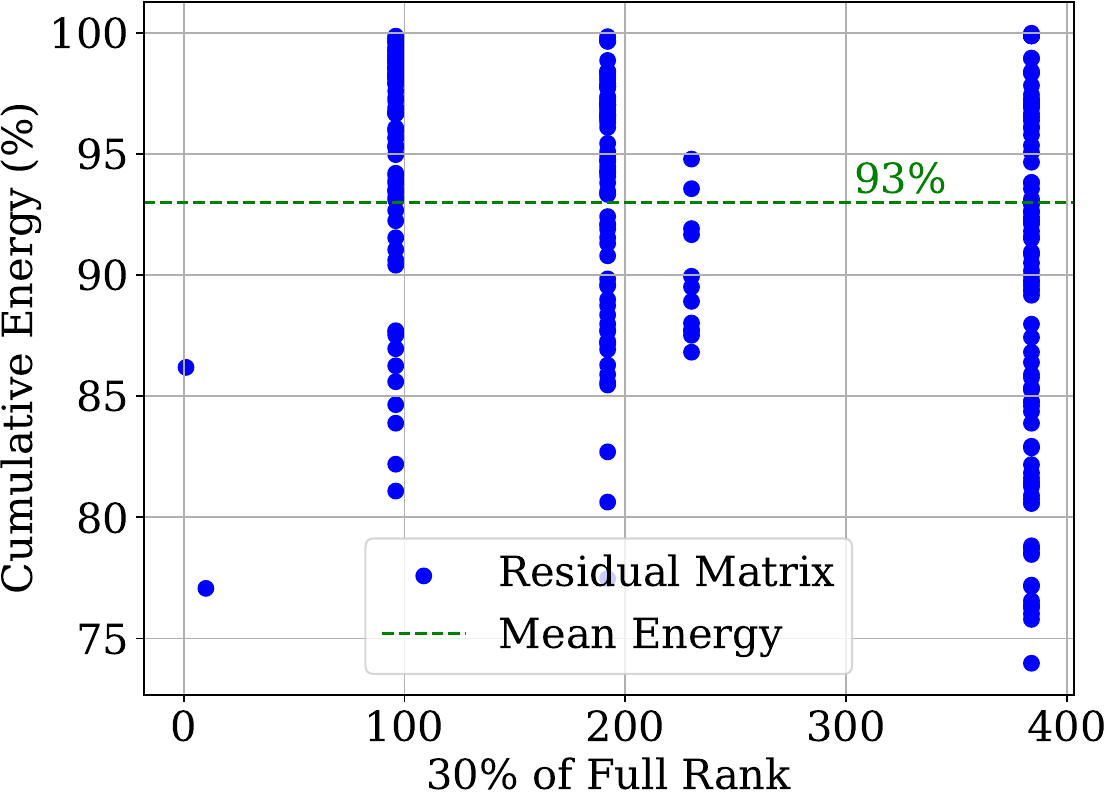}
    \vspace{-0.55cm}
    \caption{}
    \label{fig:mot-2}
  \end{subfigure}
  \begin{subfigure}[c]{0.39\linewidth}
    \centering
    \includegraphics[width=\linewidth]{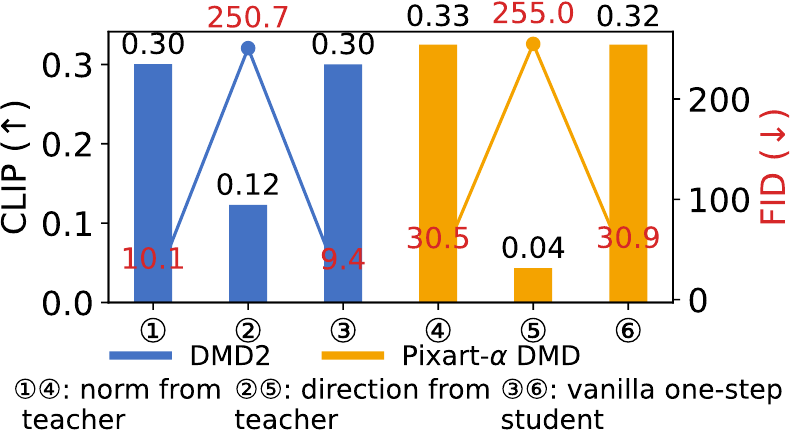}
    \vspace{-0.6cm}
    \caption{}
    \label{fig:mot-3}
  \end{subfigure}
  \begin{subfigure}[c]{0.32\linewidth}
    \centering
    \includegraphics[width=\linewidth]{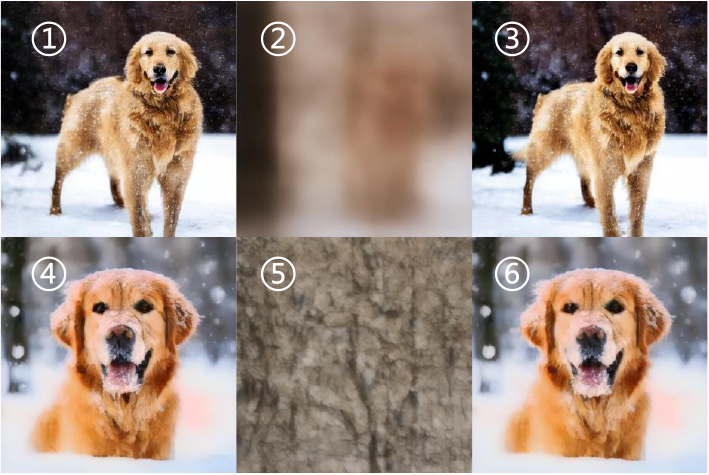}
    \vspace{-0.6cm}
    \caption{}
    \label{fig:mot-4}
  \end{subfigure}
\begin{subfigure}[c]{0.175\linewidth}
    \centering
    \includegraphics[width=\linewidth]{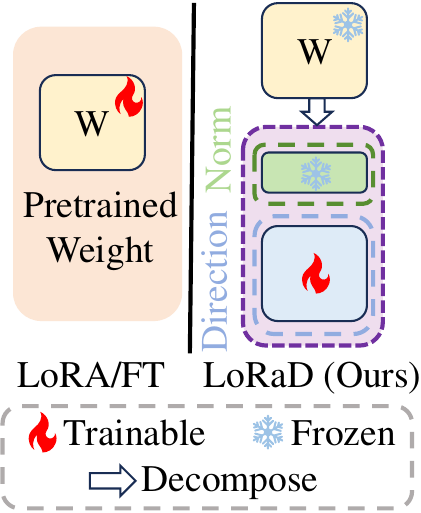}
    \vspace{-0.6cm}
    \caption{}
    \label{fig:mot-5}
  \end{subfigure}
  \vspace{-0.35cm}
  \caption{Motivational analysis of our method. (a) Differences in weight norm and direction between the one-step student and the teacher model. See \textit{\textcolor{blue}{Suppl.~E}} for details and additional examples. (b) SVD analysis of the residual matrix for DMD2. (c) Replacing the one-step model's norm with that of the multi-step model has little effect (\textcircled{1}, \textcircled{4}); replacing the direction severely degrades generation quality (\textcircled{2}, \textcircled{5}). (d) Qualitative examples corresponding to \subref{fig:mot-3}. (e) Illustration of LoRaD.} 
  \label{fig:motivation}
  \vspace{-0.5cm}
  
\end{figure*}

\section{Introduction}
\label{sec:intro}

Diffusion models (DMs)~\citep{ho2020denoising,sohl2015deep,song2020score} have received considerable attention for their ability to generate high-quality and diverse content. Thus, they are widely applied to tasks such as text-to-image~\citep{rombach2022high,li2024photomaker,ruiz2023dreambooth,zhang2023adding} generation, text-to-video~\citep{khachatryan2023text2video,wu2023tune,zhou2024storydiffusion,kong2024hunyuanvideo} generation, and image-to-video~\citep{wang2025wan,ni2023conditional,bar2024lumiere,hu2025lamd} generation.
However, the reliance of DMs on multiple sampling steps leads to high computational cost and slow inference. To address this, recent distillation methods reduce the number of steps to a few~\citep{luo2023lcm,chadebec2025flash} or even one~\citep{ren2024hyper,lin2024sdxl,dao2024swiftbrush}. Interestingly, during distillation, we find that the weight norm remains relatively small across layers, while the direction shows larger variations when reparameterizing weights into \textit{norm} and \textit{direction} for both teacher and student generators.


Inspired by the weight reparameterization~\citep{salimans2016weight,liu2024dora}, we adopt a similar decomposition to analyze weight changes in diffusion distillation. To begin our analysis, we examine weight updates between state-of-the-art (SOTA) one-step models (\textit{e.g.}, DMD2~\citep{yin2024improved} and Pixart-$\alpha$ DMD~\citep{yin2024one}) and their corresponding multi-step counterparts (\textit{e.g.}, SD 1.5~\citep{rombach2022high} and Pixart-$\alpha$~\citep{chen2023pixart}). As shown in Fig.~\ref{fig:motivation}~(\subref{fig:mot-1})~(left), in U-Net–based architectures, the weight norm remains nearly stable across layers, with a mean and standard deviation (STD) of 0.1\% and 0.2\%, respectively. In contrast, the weight direction exhibits a much more pronounced change, with a mean of 2.2\% and STD of 2.1\%, corresponding to ratios of 22× and 10× those of the norm. A similar trend is observed in DiT–based architectures (see Fig.~\ref{fig:motivation}~(\subref{fig:mot-1})~(right)). These observations suggest that the weight direction may carry richer and more sensitive information than the norm in distillation. Further, if the direction indeed accounts for the primary information differences, we ask whether these differences exhibit a structured pattern. To this end, we perform SVD on the residual matrix—the difference between the one-step and multi-step direction matrices—and find that retaining 30\% of its rank recovers 93\% of the information, highlighting its low-rank nature (see Fig.~\ref{fig:motivation}~(\subref{fig:mot-2})). 

To quantify the impact of these two components, we conduct a controlled ablation study by selectively replacing either the norm or direction of the one-step model with that from the multi-step teacher (see Fig.~\ref{fig:motivation}~(\subref{fig:mot-4})). As shown in Fig.~\ref{fig:motivation}~(\subref{fig:mot-3}), substituting the norm leads to negligible performance change (e.g., DMD2: +0.7 FID, unchanged CLIP), whereas substituting the direction causes severe degradation (e.g., DMD2: +241.3 FID, -0.18 CLIP). These findings suggest that the weight direction plays a primary role in distillation, while variation in the norm appears comparatively minor. One possible explanation is that initializing the student with teacher weights aligns the initial norm, and weight decay during training further constrains norm drift~\citep{loshchilov2017decoupled}; the distillation signal then acts mainly through adjustments in the weight direction to reduce representational discrepancy~\citep{salimans2016weight}. Taken together, these results indicate that \textbf{\textit{direction reconstruction is a key factor underlying performance improvement in distillation.}}

The distillation methods mentioned above can be broadly categorized into two types: full fine-tuning (FT) and Low-Rank Adaptation (LoRA)~\citep{hu2022lora}-based fine-tuning. However, they directly update the model parameters while optimizing both norm and direction. The changes in norm and direction differ, with norm showing minimal variation and directions experiencing significant changes, which increases the optimization difficulty due to the strong coupling between them. Furthermore, both FT and LoRA face issues of slow convergence~\citep{huang2024lazy,dong2024fine}, instability~\citep{han2024parameter,hayou2024impact}, and overfitting~\citep{aghajanyan2020intrinsic,huang2025comlora}, further complicating the optimization process.

To address the above challenges, we propose Low-rank Rotation of weight Direction (LoRaD) (see Fig.~\ref{fig:motivation}~(\subref{fig:mot-5})), which adjusts the direction of pre-trained weights via learnable rotation matrices. Given the structured nature (\textit{i.e.}, low-rank property) of directional changes, the rotation angles are parameterized as the product of two low-rank matrices to further reduce the number of learnable parameters. We integrate LoRaD into Variational Score Distillation (VSD)~\citep{wang2023prolificdreamer} and introduce Weight Direction-aware Distillation (WaDi), a novel one-step text-to-image distillation framework. Experiments on the COCO 2014~\citep{lin2014microsoft} and COCO 2017~\citep{lin2014microsoft} datasets show that WaDi achieves SOTA FID scores, outperforming all existing one-step generation methods. This was accomplished by optimizing only the direction, which reduced the difficulty of distillation, while using only about \textbf{10\%} of the U-Net parameters as trainable components—greatly enhancing parameter efficiency. Furthermore, we apply WaDi to downstream tasks including controllable generation, relation inversion, high-resolution synthesis, and image customization, demonstrating its acceleration capability and broad applicability. Our contributions are summarized as follows:
\begin{itemize}[leftmargin=*,topsep=0pt, itemsep=4pt, parsep=0pt]
    \item We conduct an in-depth analysis of weight changes in U-Net between multi-step and one-step generation models, which points to weight-direction adjustment as a key driver of one-step distillation. This provides a new theoretical perspective for efficient distillation.
    \item We propose a novel distillation framework for one-step text-to-image generation, named WaDi, which employs LoRaD to model weight directions via low-rank rotations, effectively guiding the student model to align with the teacher distribution.
    \item WaDi is evaluated on the COCO dataset and several downstream tasks. Both qualitative and quantitative results demonstrate that WaDi significantly improves inference efficiency while achieving substantial gains in image quality. 
\end{itemize}

%% file: sec/2_related_work.tex
\section{Related Work}
\label{gen_inst}

\minisection{Diffusion models.}
Diffusion models~\citep{ho2020denoising,sohl2015deep,song2019generative,song2020score,hu2025adaptive,hu2025exploiting,gao2025towards,wu2025ret3d} excel in image generation, but pixel-space computation imposes a heavy computational burden. To improve efficiency, \citet{rombach2022high} introduced Latent Diffusion Models (LDM), shifting denoising to latent space. However, existing text-guided methods~\citep{rombach2022high,podell2023sdxl,li2024photomaker,ruiz2023dreambooth,zhang2023adding} are still slow due to multi-step generation. While most use a U-Net backbone, Diffusion Transformer (DiT)~\citep{peebles2023scalable} replaces it with a Transformer for better scalability, advancing text-to-image generation~\citep{chen2023pixart,chen2024pixart,chen2024pixart-sigma,esser2024scaling}. Despite improvements, iterative denoising remains a slow process. Recently, many acceleration methods have emerged.

\minisection{Diffusion model acceleration.}
The existing acceleration methods can be divided into training-free and training-based approaches. \textit{Training-free acceleration methods} for diffusion models fall into two main categories. The first method, which reduces redundant computation through caching~\citep{ma2024deepcache,wimbauer2024cache,selvaraju2024fora,li2024faster}, is exemplified by Faster Diffusion~\citep{li2024faster}. The second method uses high-order solvers~\citep{song2020denoising,liu2022pseudo,zhang2022fast,lu2022dpm,lu2022dpm++}, such as DDIM~\citep{song2020denoising} and DPM-Solver~\citep{lu2022dpm,lu2022dpm++}, to reduce the number of sampling steps. However, the acceleration effects of these two methods are limited, so training-based methods have received more attention. 

\textit{Training-based acceleration methods} can be broadly categorized into four groups: consistency distillation (CD), progressive distillation (PD), diffusion-GAN distillation, and variational score distillation (VSD). CD~\citep{song2023consistency,wang2024phased,ren2024hyper,kim2023consistency,luo2023latent,luo2023lcm} learns trajectory-level consistency for faster sampling but often suffers from low image fidelity. PD~\citep{salimans2022progressive,ren2024hyper} reduces steps in stages, introducing significant training overhead. Diffusion-GAN distillation~\citep{luo2024you,lin2024sdxl,xu2024ufogen,kang2024distilling}, such as Diffusion2GAN~\citep{kang2024distilling}, enhances fidelity by distilling multi-step diffusion into a GAN. VSD adopts a dual-teacher strategy for distribution alignment~\citep{dao2024swiftbrush,nguyen2024swiftbrush,zhou2024long,yin2024improved,yin2024one}. SwiftBrush~\citep{nguyen2024swiftbrush} achieves one-step, image-free generation. SwiftBrushv2~\citep{dao2024swiftbrush} leverages model ensembling, while DMD~\citep{yin2024one} employs a regression loss to further improve performance. DMD2~\citep{yin2024improved} extends VSD to few-step generation and underpins recent text-to-video acceleration frameworks~\citep{yi2025magic,shao2025magicdistillation}.

However, existing training-based methods commonly use FT or LoRA, which can increase optimization difficulty. We find that directional changes are generally more influential in distillation. Therefore, we propose WaDi, which leverages LoRaD to focus on modeling directional rotations.

%% file: sec/3_method.tex
\section{Method}
\label{headings}
\begin{figure*}[t]
  \centering

    \includegraphics[width=0.8\linewidth]{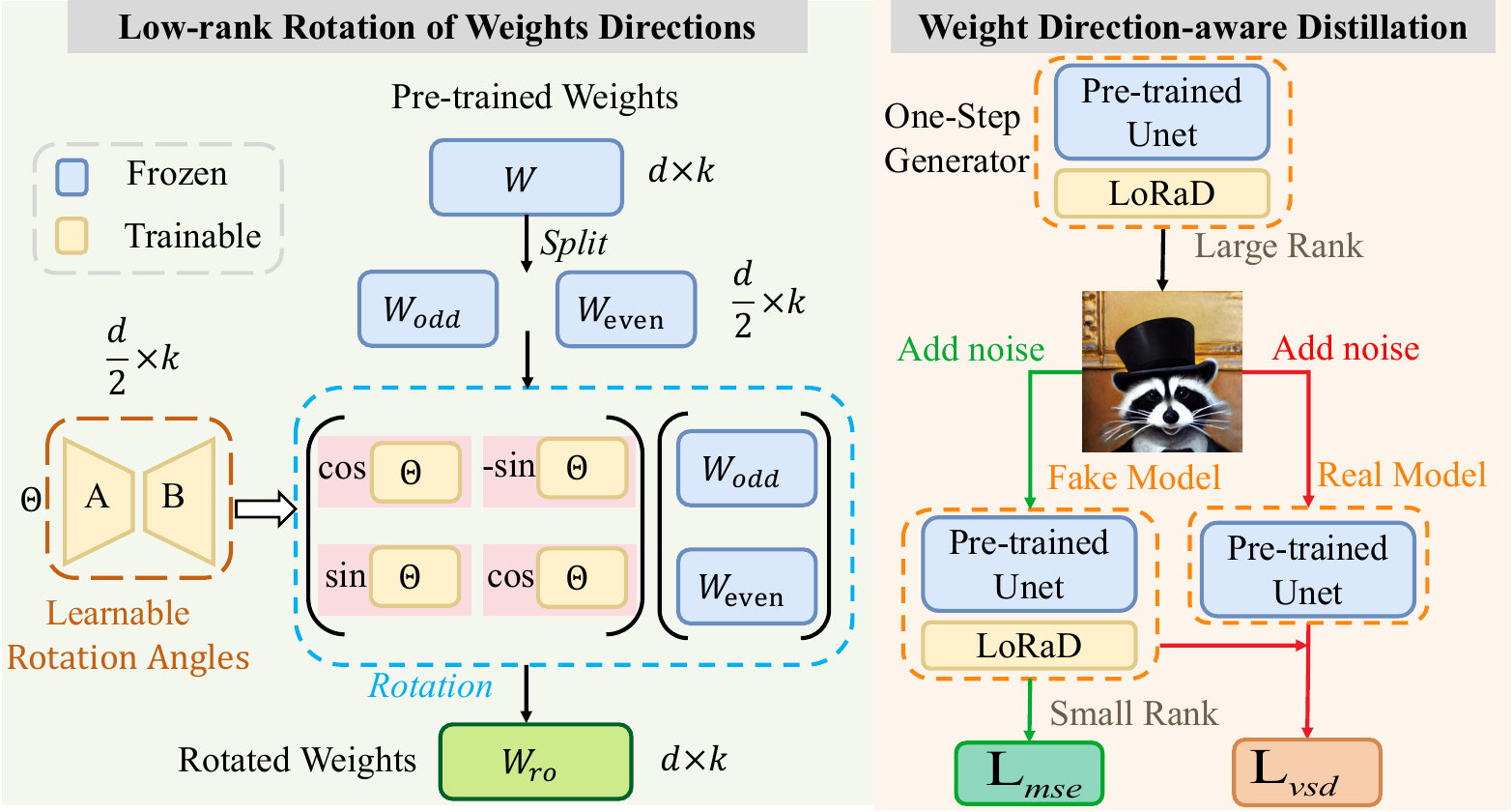}
\vspace{-0.3cm}
  \caption{(Left) Detailed architecture of the Low-rank Rotation of weight Direction (LoRaD) module. The LoRaD rotates the pre-trained weight directions using learnable low-rank rotation angles. (Right) Overview of the Weight Direction-aware Distillation (WaDi) framework. }
  \label{fig:dkd}
\vspace{-0.6cm}
\end{figure*}

We first provide a brief overview of Variational Score Distillation (VSD) in Section~\ref{sec:background}, which serves as the foundation of our work. Motivated by the observation that weight direction changes play a key role in distillation, we introduce a \textit{Low-rank Rotation of weight Direction} (LoRaD) module in Section~\ref{sec:lorad} (See \textit{\textcolor{blue}{Suppl.~D}} for more theoretical explanation.). Finally, we integrate LoRaD into the VSD to form our proposed distillation framework, \textit{Weight Direction-aware Distillation} (WaDi).

\subsection{Preliminary}
\label{sec:background}
\textbf{Latent Diffusion Models} (LDM)~\citep{rombach2022high} perform the diffusion process in a low-dimensional latent space, which improves computational efficiency. The training objective of LDM can be formulated as:
\begin{equation}
    \mathcal{L}_{mse}=\min_{\varphi} \mathbb{E}_{t, \epsilon,\boldsymbol{c}}\left\|\epsilon_\varphi\left(\boldsymbol{z}_t,\boldsymbol{c} ,t\right)-\epsilon\right\|_2^2,
    \label{equ:ldm}
\end{equation}
where $\epsilon \sim \mathcal{N}(0, I)$ is Gaussian noise, $\boldsymbol{z}_t$ is the latent variable at timestep $t$, and $\boldsymbol{c}$ denotes the condition (\textit{e.g.}, prompt) used to guide image generation. $\epsilon_\varphi\left(\boldsymbol{z}_t, \boldsymbol{c},t\right)$ is the noise predicted by the model parameterized by $\varphi$.

\textbf{Variational Score Distillation} (VSD)~\citep{wang2023prolificdreamer} was initially proposed for text-to-3D generation to address issues such as oversaturation and reduced diversity. It was subsequently extended to 2D text-to-image generation in methods such as Swiftbrush~\citep{nguyen2024swiftbrush}, DMD~\citep{yin2024one,yin2024improved}, and SiD~\citep{zhou2024score,zhou2024long}, enabling one-step generation. The training objective of VSD is formulated as:
\begin{equation}
\begin{aligned}
\nabla_\lambda \mathcal{L}_{\mathrm{vsd}} &=
\mathbb{E}_{t,\epsilon,\boldsymbol{c}} \Bigl[
\omega(t)\left(
\epsilon_\psi(\boldsymbol{z}_t,\boldsymbol{c},t)
\right.
\\ &\qquad
\left.-\,\epsilon_\phi(\boldsymbol{z}_t,\boldsymbol{c},t)
\right)
\frac{\partial G_\lambda(\boldsymbol{z}_{\mathrm{init}},\boldsymbol{c})}{\partial \lambda}
\Bigr].
\label{equ:vsd}
\end{aligned}
\end{equation}

where $\omega(t)$ is a time-dependent weighting term, $\epsilon_\psi$ is the real model parameterized by $\psi$, $\epsilon_\phi$ is the fake model parameterized by $\phi$, and $G_\lambda$ is the one-step generator parameterized by $\lambda$, with $\boldsymbol{z}_{init} \sim \mathcal{N}(0, I)$ as its input noise. Additionally, $\epsilon_\phi$ is trained using Eq.~(\ref{equ:ldm}). VSD alternates between updating $\epsilon_\phi$ and $G_\lambda$ until convergence.

\subsection{Low-rank Rotation of Weight Direction}
\label{sec:lorad}
Analyzing the weight changes between multi-step U-Net models and their one-step counterparts suggests notable directional shifts with relatively small changes in norm. Motivated by this, we propose Low-rank Rotation of weight Direction (LoRaD) (see Fig.~\ref{fig:dkd}~(left)), which updates weights by learning rotations that alter only their directions. Furthermore, we observe that the changes in weight direction exhibit a low-rank structure (see Fig.~\ref{fig:motivation}~(\subref{fig:mot-2})). To exploit this property and reduce the overhead of full-rank modeling, which introduces additional parameters equivalent to 50\% of the original weights, we adopt the low-rank decomposition strategy of LoRA~\citep{hu2022lora}.
Starting from the 2D case ($d=2$), given a weight vector $\alpha \in \mathbb{R}^{d}$, we apply a 2D rotation matrix as follows:
\begin{equation}
    \alpha_{ro}=\left(\begin{array}{cc}
\cos \theta & -\sin \theta \\
\sin \theta & \cos \theta
\end{array}\right)\binom{\alpha^{(1)}}{\alpha^{(2)}},
\end{equation}
where $\alpha_{ro}$ is the rotated weight vector.
Inspired by the Rotary Position Embedding (RoPE)~\citep{su2024roformer}, which generalizes the 2D case to any even dimension $d$, we apply a different rotation matrix\footnote{We do not need to explicitly separate the norm matrix, as rotations do not affect norm.} to each column of the pre-trained weight matrix $W \in \mathbb{R}^{d \times k}$:
\begin{equation}
    W_{ro}=\left[R_{\Theta,1}^d W_{\cdot,1}, R_{\Theta,2}^d W_{\cdot,2}, \cdots, R_{\Theta,k}^d W_{\cdot,k}\right],
\end{equation}
where the rotation matrices $R_{\Theta}=\{R_{\Theta,i}^d\}_{i=1}^k$ are defined as:
\begingroup
\setlength{\arraycolsep}{2pt}   
\renewcommand{\arraystretch}{0.95}
\begin{equation}
\resizebox{\columnwidth}{!}{$
R_{\Theta,i}^d=
\left(\begin{array}{@{}ccccccc@{}}
\cos\theta_{1,i}&-\sin\theta_{1,i}&0&0&\cdots&0&0\\
\sin\theta_{1,i}& \cos\theta_{1,i}&0&0&\cdots&0&0\\
0&0&\cos\theta_{2,i}&-\sin\theta_{2,i}&\cdots&0&0\\
0&0&\sin\theta_{2,i}& \cos\theta_{2,i}&\cdots&0&0\\
\vdots&\vdots&\vdots&\vdots&\ddots&\vdots&\vdots\\
0&0&0&0&\cdots&\cos\theta_{\frac d2,i}&-\sin\theta_{\frac d2,i}\\
0&0&0&0&\cdots&\sin\theta_{\frac d2,i}& \cos\theta_{\frac d2,i}
\end{array}\right),
$}
\label{equ:roma}
\end{equation}
\endgroup
where $\Theta=\left\{\theta_j\right\}_{j=1}^{\frac{d}{2}} \in \mathbb{R}^{\frac{d}{2} \times k}$.

Given the sparsity of $R_{\Theta, i}^d$ in Eq.~(\ref{equ:roma}), the matrix-vector multiplication $R_{\Theta, i}^d W_{\cdot,i} \in \mathbb{R}^d$ can be computed efficiently as:
\begingroup
\setlength{\arraycolsep}{2pt}      
\renewcommand{\arraystretch}{0.95} 
\begin{equation}
\resizebox{\columnwidth}{!}{$
R_{\Theta,i}^{d}\, W_{\cdot,i} =
\left(\begin{array}{@{}c@{}}
W_{\cdot,i}^{(1)}\\
W_{\cdot,i}^{(2)}\\
W_{\cdot,i}^{(3)}\\
W_{\cdot,i}^{(4)}\\
\vdots\\
W_{\cdot,i}^{(d-1)}\\
W_{\cdot,i}^{(d)}
\end{array}\right)
\odot
\left(\begin{array}{@{}c@{}}
\cos\theta_{1,i}\\
\cos\theta_{1,i}\\
\cos\theta_{2,i}\\
\cos\theta_{2,i}\\
\vdots\\
\cos\theta_{\frac d2,i}\\
\cos\theta_{\frac d2,i}
\end{array}\right)
+
\left(\begin{array}{@{}c@{}}
W_{\cdot,i}^{(1)}\\
W_{\cdot,i}^{(2)}\\
W_{\cdot,i}^{(3)}\\
W_{\cdot,i}^{(4)}\\
\vdots\\
W_{\cdot,i}^{(d-1)}\\
W_{\cdot,i}^{(d)}
\end{array}\right)
\odot
\left(\begin{array}{@{}c@{}}
-\sin\theta_{1,i}\\
\ \sin\theta_{1,i}\\
-\sin\theta_{2,i}\\
\ \sin\theta_{2,i}\\
\vdots\\
-\sin\theta_{\frac d2,i}\\
\ \sin\theta_{\frac d2,i}
\end{array}\right),
$}
\label{equ:equ_6}
\end{equation}
\endgroup
where $\odot$ denotes element-wise multiplication. This implementation leverages the sparsity of the rotation matrix, allowing the computation to be performed using only element-wise operations, thus significantly reducing the computational cost.

Furthermore, since the rotation matrices in Eqs.~(\ref{equ:roma}) and~(\ref{equ:equ_6}) are block-diagonal with independent $2 \times 2$ submatrices, the computation can be efficiently implemented as a parallel application of multiple $2\times2$ rotations across odd-even index pairs. As shown in Fig.~\ref{fig:dkd} (left), we split the $d$-dimensional space of the pre-trained weight matrix $W \in \mathbb{R}^{d \times k}$ into $\frac{d}{2}$ subspaces and rotate each independently. By separating the odd and even rows of $W$, we define:
\begin{equation}
    \begin{gathered}
W_{\text {odd }}=\left(W^{(1)}, W^{(3)}, \ldots, W^{(d-1)}\right)^T, \\
W_{\text {even }}=\left(W^{(2)}, W^{(4)}, \ldots, W_{\cdot, i}^{(d)}\right)^T,
\end{gathered}
\end{equation}
resulting in two matrices $W_{\text {odd }} \in \mathbb{R}^{\frac{d}{2} \times k}$ and $W_{\text {even }} \in \mathbb{R}^{\frac{d}{2} \times k}$.

The resulting parallel $2 \times 2$ rotations over each odd-even row pair can be expressed compactly as:
\begin{equation}
    W_{ro} = R_{\Theta}W=\left[\begin{array}{cc}
\cos \Theta & -\sin \Theta \\
\sin \Theta & \cos \Theta
\end{array}\right]\left[\begin{array}{l}
W_{\text {odd }} \\
W_{\text {even }}
\end{array}\right],
\label{equ:rotary}
\end{equation}
where $W_{ro} \in \mathbb{R}^{d \times k}$ is the rotated weight matrix, and $\Theta \in \mathbb{R}^{\frac{d}{2} \times k}$ is the learnable rotation angle parameter matrix. To further reduce the number of trainable parameters, we apply low-rank decomposition to $\Theta$, inspired by LoRA~\citep{hu2022lora}, as follows:
\begin{equation}
    \Theta = A B,
\end{equation}
where $A \in \mathbb{R}^{\frac{d}{2} \times r}$ and $B \in \mathbb{R}^{r \times k}$ are low-rank parameter matrices with rank $r$. Finally, Eq.~(\ref{equ:rotary}) can be rewritten as:
\begingroup
\setlength{\arraycolsep}{2pt}      
\renewcommand{\arraystretch}{0.95} 
\begin{equation}
\resizebox{\columnwidth}{!}{$
W_{ro}=R_{\Theta}W=R_{AB}W=
\left[\begin{array}{@{}cc@{}}
\cos AB & -\sin AB\\
\sin AB & \ \cos AB
\end{array}\right]
\left[\begin{array}{@{}c@{}}
W_{\text{odd}}\\
W_{\text{even}}
\end{array}\right].
$}
\end{equation}
\endgroup

\subsection{Weight Direction-aware Distillation}
\label{sec:dkd}
To fully leverage the directional characteristics observed in distillation, we integrate LoRaD into the VSD. This yields a direction-aware distillation framework, which we term Weight Direction-aware Distillation (WaDi).
As illustrated in Fig.~\ref{fig:dkd} (right), WaDi employs a pre-trained diffusion model $\epsilon_\psi$ as the teacher (real model) and introduces a trainable fake model $\epsilon_\phi$ (initialized from $\epsilon_\psi$) to approximate the teacher's distribution. The final student model (one-step generator) $G_{\lambda}$, also initialized from $\epsilon_\psi$, is trained to synthesize high-quality images in one-step. See \textit{\textcolor{blue}{Suppl.~F.3}} for algorithm details.

To enhance alignment with the real distribution, we apply LoRaD to both the student and fake models. Specifically, the one-step generator $G_{\lambda_{\Theta^l}}$ incorporates a high-rank rotation matrix $\Theta^l$ to better fit the teacher, while the fake model $\epsilon_{\phi_{\Theta^s}}$ uses a low-rank rotation matrix $\Theta^s$ to provide adaptive guidance. Finally, we alternate the optimization of $\lambda_{\Theta^l}$ and $\phi_{\Theta^s}$ to jointly improve the quality of the generation.

Accordingly, the WaDi training objective can be rewritten from Eq.~(\ref{equ:vsd}) as:
\begin{equation}
\begin{aligned}
\nabla_{\lambda_{\Theta^l}} \mathcal{L}_{\mathrm{wadi}}
&= \mathbb{E}_{t,\epsilon,\boldsymbol{c}}\Bigl[
\omega(t)\bigl(
\epsilon_\psi(\boldsymbol{z}_t,\boldsymbol{c},t)
\\ &\qquad
-\,\epsilon_{\phi_{\Theta^s}}(\boldsymbol{z}_t,\boldsymbol{c},t)
\bigr)\,
\frac{\partial G_{\lambda_{\Theta^l}}(\boldsymbol{z}_{\mathrm{init}},\boldsymbol{c})}{\partial \lambda_{\Theta^l}}
\Bigr],
\end{aligned}
\end{equation}

The training objective for $\epsilon_{\phi_{\Theta^s}}$ can also be rewritten from Eq.~(\ref{equ:ldm}) as:
\begin{equation}
    \min_{\phi_{\Theta^s}} \mathbb{E}_{t, \epsilon,\boldsymbol{c}}\left\|\epsilon_{\phi_{\Theta^s}}\left(\boldsymbol{z}_t,\boldsymbol{c} ,t\right)-\epsilon\right\|_2^2.
\end{equation}

%% file: sec/4_experiment.tex
\begin{table*}[t]
\centering
\caption{Quantitative comparison of WaDi and other methods on zero-shot COCO 2014 results. $^\ast$ indicates our reproduced results, and $^\wr$ indicates results using the official pre-trained models. ‘-’ denotes unknown. Best and second-best scores are in \textbf{bold} and \underline{underline}, respectively. ``Image-free" refers to training without supervision from real images.}
\vspace{-0.4cm}
\resizebox{\textwidth}{!}{
\begin{tabular}{cccccccccccc}
\toprule
\textbf{Method} & \textbf{\#Params} & \textbf{NFEs} & \textbf{Type} & \textbf{Trainable params} & \textbf{FID} $\downarrow$ & \textbf{CLIP} $\uparrow$ & \textbf{Precision} $\uparrow$ & \textbf{Recall} $\uparrow$ & \textbf{Image-free?} & \textbf{Training Data} \\
\midrule
\rowcolor{gray!20} \multicolumn{11}{c}{Stable Diffusion 1.5-based backbone} \\
SD 1.5 ($\textit{cfg}=3.0$) & 860M & 25 & U-Net & 860M & 8.78 & 0.30 & 0.59 & 0.53 & \xmark & 5B \\
\cdashline{1-11}
LCM-LoRA$^\wr$ & 860M & 1 & LoRA & 67.50M & 77.73 & 0.24 & 0.22 & 0.15 & \xmark & 12M \\
InstaFlow & 860M & 1 & U-Net & 860M & 13.10 & 0.28 & 0.53 & 0.45 & \xmark & 3.2M \\
UFOGen & 860M & 1 & U-Net & 860M & 12.78 & - & - & - & \xmark & 12M \\
DMD & 860M & 1 & U-Net & 860M & \underline{11.49} & \textbf{0.32} & - & - & \xmark & 3M \\
DMD2$^\ast$ & 860M & 1 & U-Net & 860M & 12.96 & 0.30& \underline{0.60} & \underline{0.47} & \cmark & 1.4M \\
SiD-LSG$^\ast$ & 860M & 1 & U-Net & 860M & 14.27 & 0.30& 0.56 & \textbf{0.48} & \cmark & 1.4M \\
PCM & 860M & 1 & U-Net & 860M & 17.91 & 0.29& - & - & \xmark & 3M \\
Hyper-SD$^\wr$ & 860M & 1 & LoRA & 67.25M & 22.90 & \underline{0.31}& \textbf{0.62} & 0.25 & \xmark & - \\
YOSO$^\wr$ & 860M & 1 & LoRA & 67.25M & 23.68 & 0.29 & 0.56 & 0.36 & \xmark & 4M \\
\rowcolor{blue!10} WaDi & 860M & 1 & LoRaD & 83.80M & \textbf{10.79} & \underline{0.31} & \textbf{0.62} & \textbf{0.48} & \cmark & 1.4M \\
\midrule
\rowcolor{gray!20} \multicolumn{11}{c}{Stable Diffusion 2.1-based backbone} \\
SD 2.1 ($\textit{cfg}=3.0$) & 865M & 1 & U-Net & 865M & 9.60 & 0.32 & 0.59 & 0.50 & \xmark & 5B \\
\cdashline{1-11}
SD-Turbo$^\wr$ & 865M & 1 & U-Net & 865M & 16.14 & \textbf{0.33} & \textbf{0.65} & 0.35 & \xmark & - \\
Swiftbrush & 865M & 1 & U-Net & 865M & 16.67 & 0.29 & 0.47 & 0.46 & \cmark & 1.4M \\
Swiftbrushv2$^\ast$ & 865M & 1 & U-Net+LoRA & 884.14M & 15.98 & \textbf{0.33} & 0.58 & \underline{0.47} & \cmark & 1.4M \\
SiD-LSG$^\ast$ & 865M & 1 & U-Net & 865M & 15.17 & 0.30 & 0.56 & 0.46 & \cmark & 1.4M \\
TiUE$^\wr$ & 865M & 1 & U-Net & 865M & \underline{13.49} & \underline{0.31} & 0.59 & \textbf{0.48} & \cmark & 1.4M \\
\rowcolor{blue!10} WaDi & 865M & 1 & LoRaD & 94.43M & \textbf{12.34} & \underline{0.31} & \underline{0.60} & \textbf{0.48} & \cmark & 1.4M \\
\midrule 
\rowcolor{gray!20} \multicolumn{11}{c}{PixArt-$\alpha$-based backbone} \\
PixArt-$\alpha$ ($\textit{cfg}=4.5$)$^\wr$ & 610.86M & 20 & DiT & 610.86M & 8.75 & 0.32 & 0.75 & 0.45 & \xmark & 25M \\
\cdashline{1-11}
Swiftbrush$^\ast$ & 610.86M & 1 & DiT & 610.86M & 29.89 & \underline{0.28} & 0.50 & 0.26 & \cmark & 1.4M \\
PG-SB$^\ast$ & 610.86M & 1 & DiT & 610.86M & \underline{25.58} & \underline{0.28} & \underline{0.53} & \underline{0.27} & \cmark & 1.4M \\
\rowcolor{blue!10} WaDi & 610.86M & 1 & LoRaD & 81.22M & \textbf{18.99} & \textbf{0.30} & \textbf{0.64} & \textbf{0.29} & \cmark & 1.4M \\
\bottomrule
\end{tabular}}
\label{tab:main_score}
\vspace{-0.5cm}
\end{table*}

\section{Experiment}
\label{others}
\subsection{Experimental Setup}
\label{sec:imple}

\minisection{Evaluation Datasets and Metrics.}
We systematically evaluate the zero-shot text-to-image generation capability of WaDi on the COCO 2014~\citep{lin2014microsoft} and COCO 2017~\citep{lin2014microsoft} datasets, using 30k and 5k randomly sampled images, respectively. To comprehensively assess the quality of the generation, we use the Fréchet Inception Distance (FID)~\citep{heusel2017gans} to measure image fidelity and the CLIP score~\citep{radford2021learning} to evaluate the semantic alignment of text-image. The FID is calculated using Inception V3~\citep{szegedy2016rethinking} as the feature extractor, while the CLIP score is based on the ViT-G/14~\citep{cherti2023reproducible} model. We further adopt precision and recall~\citep{kynkaanniemi2019improved} to evaluate fidelity and diversity. Finally, we also evaluate text-image alignment on the Human Preference Score v2 (HPSv2)~\citep{wu2023human} benchmark. See \textit{\textcolor{blue}{Suppl.~G.1}} for details.

\minisection{Implementation Details.} Following prior methods~\citep{nguyen2024swiftbrush,dao2024swiftbrush,yin2024improved,yin2024one}, the student model in WaDi adopts the same architecture as the teacher and is initialized with the teacher’s weights.
WaDi is trained on 1.4 M prompts sampled from the JourneyDB~\citep{sun2023journeydb} dataset. During training, the learning rate (LR) for the student is set to 1$e$-4, while the fake model uses an LR of 1$e$-2. We use AdamW~\citep{loshchilovdecoupled} as the optimizer, with a batch size of 128 (16 per GPU). The classifier-free guidance (CFG) scale is set to 1.5, and the training is conducted for 2 epochs. We distill student models based on three different backbones, namely SD 1.5~\citep{rombach2022high}, SD 2.1~\citep{rombach2022high}, and PixArt-$\alpha$ ($256 \times 256$)~\citep{chen2023pixart}. For SD 1.5 and SD 2.1, the LoRaD rank of the student is set to 256, while for PixArt-$\alpha$, it is set to 128. The LoRaD rank for all fake models is uniformly set to 32. See \textit{\textcolor{blue}{Suppl.~F.1}} for details.

\subsection{Comparison with State-of-the-Art Methods}

\minisection{Quantitative results.} We comprehensively evaluate WaDi on the COCO 2014 dataset against SOTA zero-shot one-step generation methods across three backbones: SD 1.5, SD 2.1, and PixArt-$\alpha$. To ensure fair comparison and considering computational constraints, we follow the setup of TiUE~\citep{li2025one} and uniformly reproduce WaDi, DMD2, SiD-LSG, and SwiftBrushv2 using 1.4M prompts. As shown in Tab.~\ref{tab:main_score}, WaDi achieves the best FID and Recall scores on all backbones, demonstrating superior fidelity and diversity. It also ranks first or second in CLIP and Precision, indicating strong text-image alignment and perceptual quality. Notably, only 9.74\%, 10.92\%, and 13.30\% of the model parameters are trainable for SD 1.5, SD 2.1, and PixArt-$\alpha$, respectively, highlighting WaDi’s parameter efficiency. These improvements stem from our proposed LoRaD, which reparameterizes weight updates via low-rank rotations to enable stable and efficient distillation. See \textit{\textcolor{blue}{Suppl.~F.4, G.3}}.

\minisection{Qualitative results.}
\begin{figure*}[t]
  \centering

    \includegraphics[width=0.88\linewidth]{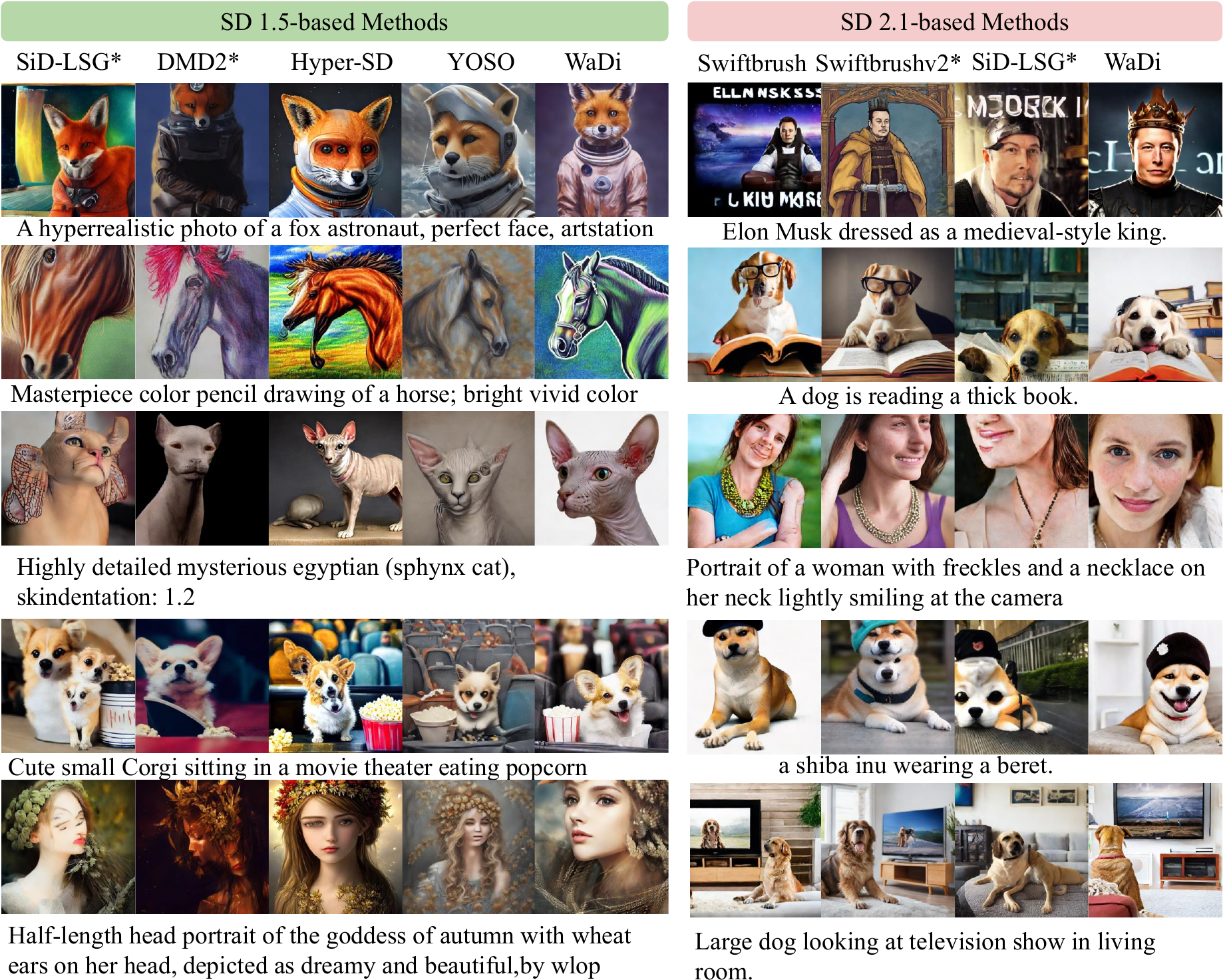}
\vspace{-0.3cm}
  \caption{Qualitative comparison with other methods, where $^\ast$ indicates our reproduced results.}
  \label{fig:compared_methods}
\vspace{-0.7cm}
\end{figure*}
Fig.~\ref{fig:compared_methods} presents a qualitative comparison of WaDi with SOTA one-step generation methods based on SD 1.5 and SD 2.1 backbones. Across diverse prompts, WaDi consistently produces visually coherent and semantically aligned results. For example, in the first and second rows, WaDi better preserves structure and stylistic fidelity, capturing sharp features and vibrant colors without artifacts or distortions. In the third and fourth rows, it accurately follows prompts involving specific subjects (\textit{e.g.}, sphynx cat, corgi, shiba inu) and contexts (\textit{e.g.}, theater, clothing), while alternative methods often miss key attributes or yield unrealistic shapes. Notably, in the last row, WaDi generates complex scenes (\textit{e.g.}, dog looking at TV) with consistent spatial composition and background details, demonstrating superior holistic understanding compared to other baselines. See \textit{\textcolor{blue}{Suppl.~G.5}}.

\subsection{Downstream Tasks}
\begin{figure}[t]
  \centering

    \includegraphics[width=0.88\linewidth]{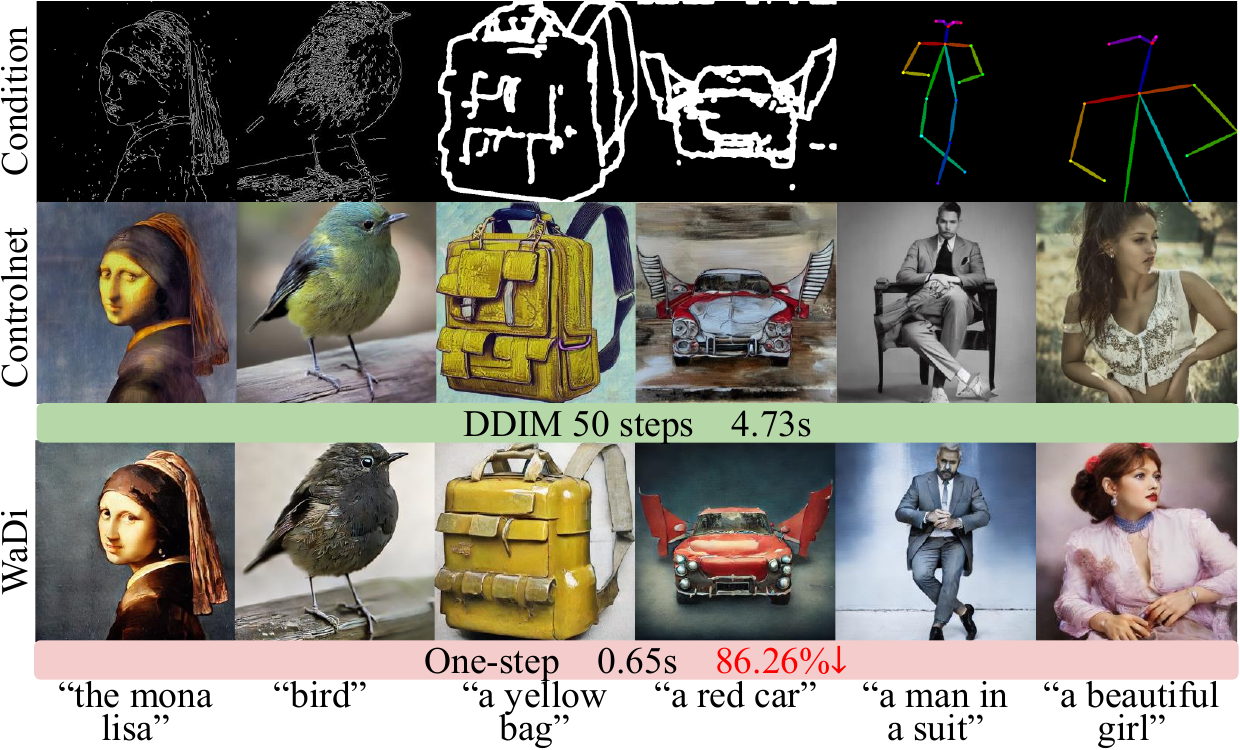}
\vspace{-0.3cm}
  \caption{Quality results by Controlnet~\citep{zhang2023adding} with or without WaDi.}
  \label{fig:controlnet}
\vspace{-0.4cm}
\end{figure}

\minisection{Controllable generation.}
ControlNet~\citep{zhang2023adding} is a widely used controllable generation model that incorporates spatial conditions into SD~\citep{rombach2022high} for fine-grained control. As shown in Fig.~\ref{fig:controlnet}, applying WaDi to ControlNet significantly improves inference efficiency, reducing inference time by \textbf{86.26\%} while preserving image quality, faithfully following spatial conditions, and maintaining prompt adherence comparable to ControlNet.

\begin{figure}[t]
  \centering

    \includegraphics[width=0.68\linewidth]{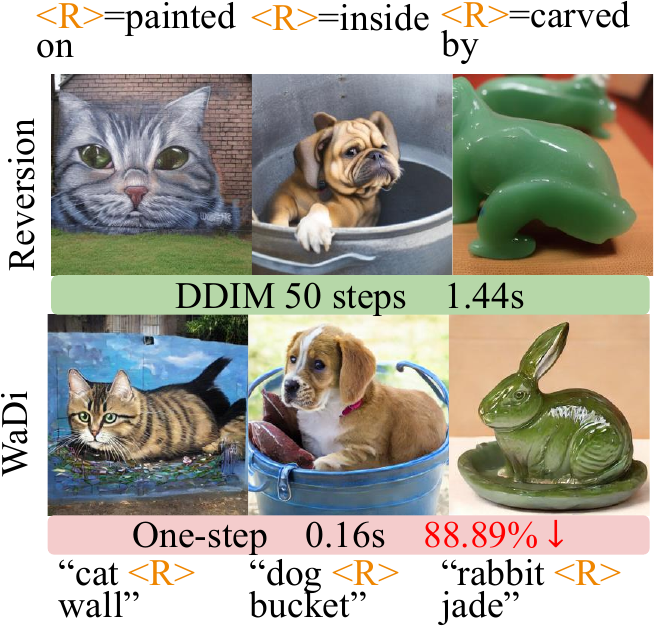}
\vspace{-0.4cm}
  \caption{Quality results by Reversion~\citep{huang2024reversion} with or without WaDi.}
  \label{fig:reversion}
\vspace{-0.8cm}
\end{figure}


\begin{figure}[t]
  \centering

    \includegraphics[width=0.68\linewidth]{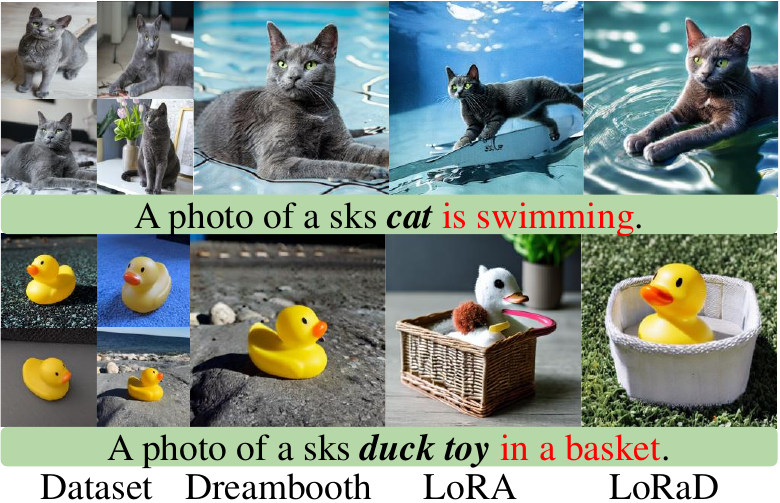}
\vspace{-0.3cm}
  \caption{Quality results by Dreambooth with or without LoRaD.}
  \label{fig:dreambooth}
\vspace{-0.4cm}
\end{figure}

\begin{table}[t]
\centering
\caption{Ablation study on the impact of adapter type in WaDi (SD 1.5, \textbf{VSD loss}) on the COCO 2017 dataset.``NM" and “DM” denote the norm mean and direction mean for all layers, respectively.}
\vspace{-0.4cm}
\resizebox{0.88\linewidth}{!}{
            \begin{tabular}{cccccc}
\toprule
 Type & \#Params & FID & CLIP & NM& DM\\
\midrule
LoRA & 120.9M & 25.27 & 0.29 & 0.06& 0.83 \\
DoRA & 121.2M & 26.56 & 0.30 & 0.03 & 0.55\\
DoRA (frozen norm) & 120.9M & 24.52 & 0.30 & - & 0.92\\
FT (DMD2) & 860.0M & 23.30 & 0.30 & 0.10& 2.21\\
LoRaD & \textbf{83.8M} & \textbf{20.86} & \textbf{0.31} &- & 2.89\\
\bottomrule
\end{tabular}}
\label{tab:abla-type}
\vspace{-0.75cm}
\end{table}

\minisection{Relation inversion.}
Reversion~\citep{huang2024reversion} is the first method to guide specific object relationship synthesis in SD via relational prompts. Integrating WaDi into Reversion significantly accelerates inference. As shown in Fig.~\ref{fig:reversion}, WaDi reduces inference time by \textbf{88.89\%} while producing high-fidelity images that align with the relational prompts, with quality close to that of the original multi-step Reversion. See \textit{\textcolor{blue}{Suppl.~F.2}} for more results.


\minisection{Image customization.}
Dreambooth~\citep{ruiz2023dreambooth} is a pioneering personalized text-to-image framework that binds the target subject to a rare token via FT of the U-Net. To enhance parameter efficiency, we integrate our proposed LoRaD into Dreambooth and compare it with Dreambooth (FT) and LoRA~\citep{hu2022lora}. As shown in Fig.~\ref{fig:dreambooth}, vanilla DreamBooth overfits by capturing the subject while memorizing training images, thus reducing prompt sensitivity. LoRA alleviates overfitting, but degrades subject identity and image fidelity. In contrast, LoRaD maintains subject fidelity while adhering to prompts, achieving a better balance. We include this DreamBooth experiment only as an illustrative example, not as a comprehensive study of diffusion fine-tuning. 

\subsection{User Study}
\label{sec:user_study}
\begin{figure}[t]
  \centering

    \includegraphics[width=\linewidth]{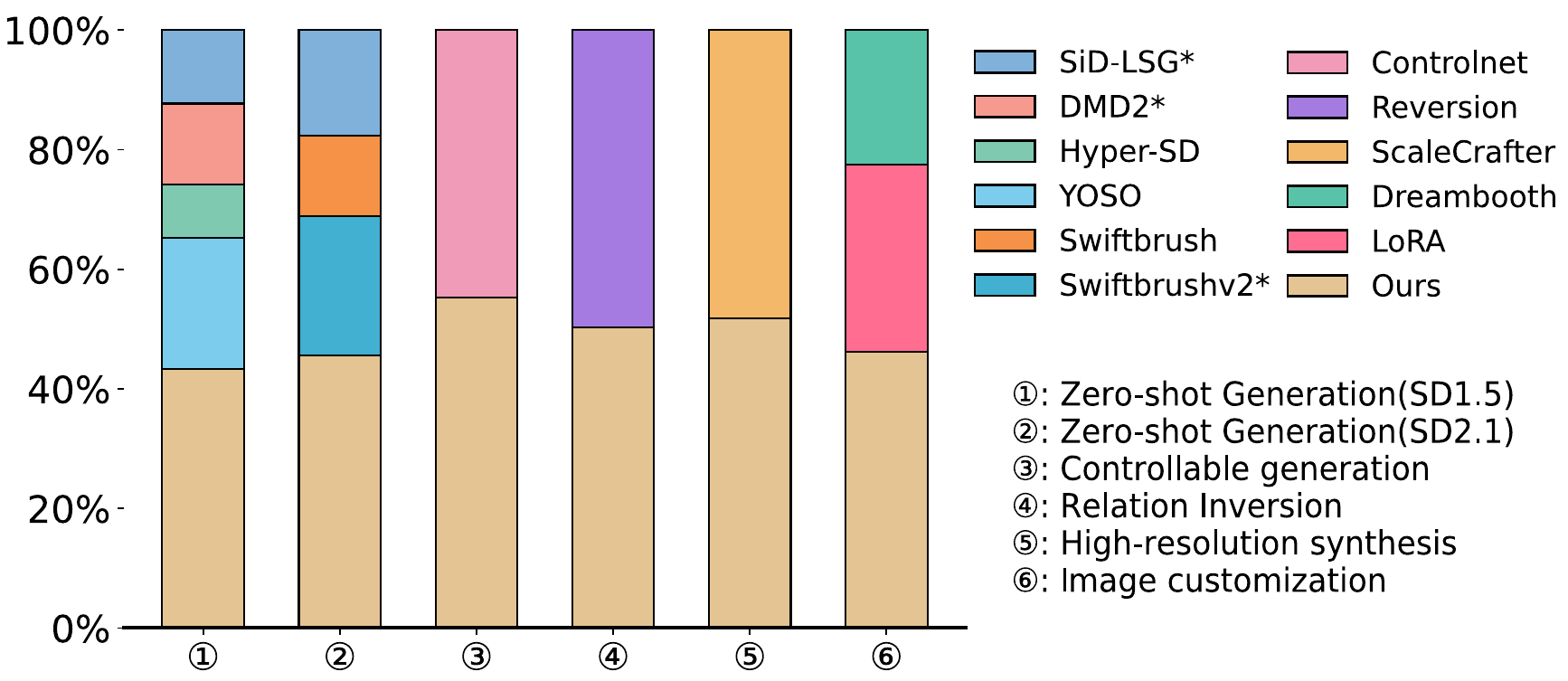}
\vspace{-0.8cm}
  \caption{User study results compared to other methods.}
  \label{fig:user_study}
\vspace{-0.65cm}
\end{figure}

To evaluate image quality and text-image alignment, we conducted a user study with 57 participants, covering zero-shot generation and downstream tasks. As shown in Fig.~\ref{fig:user_study}, the results clearly demonstrate the superiority of our method over existing baselines. See \textit{\textcolor{blue}{Suppl.~F.5}} for details.

\begin{table}[t]
\centering
\caption{Ablation study on the impact of the rank on WaDi (SD 1.5, \textbf{VSD loss}) on COCO 2014 dataset.}
\vspace{-0.4cm}
\resizebox{\linewidth}{!}{
        \begin{tabular}{ccccccc}
    \toprule
    \multirow{2}[4]{*}{Setting} & \multicolumn{4}{c}{Rank}     & \multirow{2}[4]{*}{FID} & \multirow{2}[4]{*}{CLIP} \\
\cmidrule{2-5}          & Student & \#Params & \makecell{Fake\\model} & \#Params &       &  \\
    \midrule
    A     & 64    & 20.95M & 32    & 9.38M & 13.64  & \underline{0.30}  \\
    B     & 128   & 41.90M & 32    & 9.38M & 13.16  & 0.29  \\
    C     & 256   & 83.80M & 32    & 9.38M & \textbf{10.79}  & \textbf{0.31}  \\
    D     & 512   & 167.59M & 32    & 9.38M & \underline{12.75}  & \underline{0.30}  \\
    E     & 256   & 83.80M & 16    & 4.69M & 17.53  & 0.29  \\
    F     & 256   & 83.80M & 64    & 18.76M & 16.98  & \textbf{0.31}  \\
    \bottomrule
    \end{tabular}
    }
\label{tab:ablation}
\vspace{-0.3cm}
\end{table}

\begin{figure}[t]
  \centering

    \includegraphics[width=0.68\linewidth]{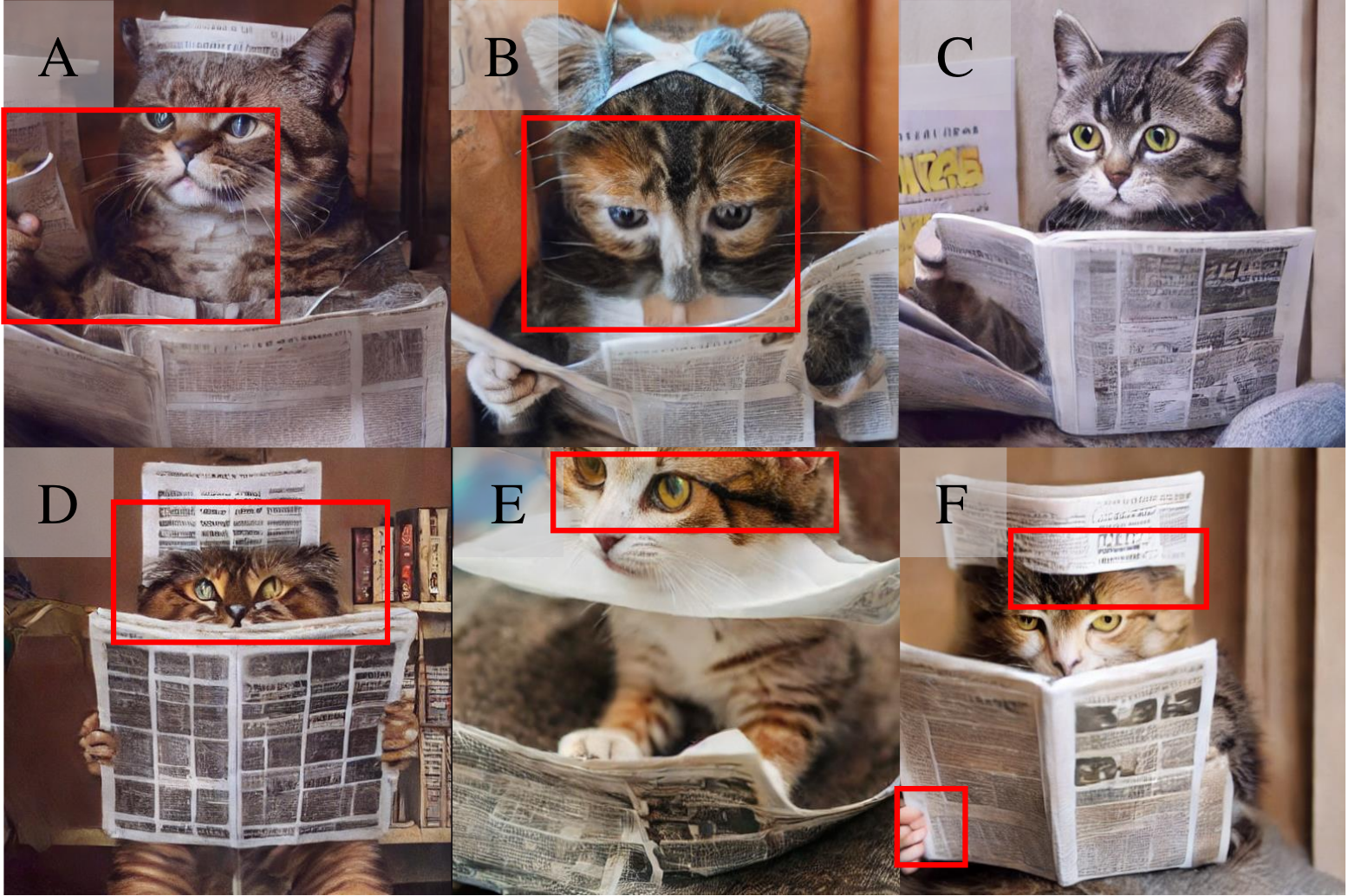}
\vspace{-0.3cm}
  \caption{One-step image generation with various settings.}
  \label{fig:abla-vis}
\vspace{-0.5cm}
\end{figure}

\subsection{Ablation Studies}

Tab.~\ref{tab:abla-type} compares five adapter types on COCO 2017 under VSD loss. LoRaD achieves the lowest FID (20.86) and competitive CLIP score (0.31) with only 83.8M trainable parameters (\verb|~|31\% fewer than LoRA/DoRA and \verb|~|90\% fewer than FT). It also yields the highest direction mean (2.89\%, vs. 2.21\% for FT and $\le$0.92\% for the LoRA/DoRA variants), indicating a broader and more effective update-direction space under a compact parameterization. Unlike DoRA and DoRA (frozen norm), which optimize directions via LoRA-style additive updates to normalized weights followed by \textbf{dynamic re-normalization}, LoRaD directly parameterizes \textbf{low-rank orthogonal rotations} of pre-trained weights, preserving norms and operating purely in direction space. Overall, LoRaD shows a favorable quality–efficiency trade-off.

We conduct an ablation study on COCO 2014 to assess the impact of rank configuration in WaDi. As shown in Tab.~\ref{tab:ablation}, we make three key observations: 1) \textbf{\textit{Increasing student rank consistently improves performance.}} Raising the rank from setting A to C reduces FID from 13.64 to 10.79, indicating that higher rank enables the student to better capture the teacher’s distribution and improve generation quality. 2) \textbf{\textit{Increasing the rank beyond a threshold yields diminishing returns.}} Comparing settings C and D, further increasing the rank degrades FID (12.75 vs. 10.79) and CLIP (0.31 vs. 0.30), suggesting that overly large ranks may cause overfitting. 3) \textbf{\textit{Fake model rank affects fidelity more than alignment.}} Varying the fake model rank (settings C, E, F) changes FID but leaves CLIP largely stable, implying fidelity is more sensitive to capacity than alignment. In summary, setting C offers a favorable trade-off between model capacity and performance, consistent with the qualitative results in Fig.~\ref{fig:abla-vis}. See \textit{\textcolor{blue}{Suppl.~G.2, G.4}} for details.


%% file: sec/5_conclusion.tex
\section{Conclusion}
This paper presents Weight Direction-aware Distillation (WaDi), an efficient one-step text-to-image distillation framework. Through an in-depth analysis of weight changes between multi-step and one-step models, we find that changes in weight direction serve as a key mechanism in distillation, while changes in norm play a comparatively smaller role. Based on this insight, we introduce the Low-rank Rotation of weight Direction (LoRaD) module to model directional adjustments in a parameter-efficient manner. Extensive experiments demonstrate that WaDi significantly outperforms existing one-step methods—such as DMD, SiD-LSG, and SwiftBrush—in both image quality and inference speed. Moreover, the distilled model can be seamlessly adapted to a wide range of downstream tasks, showcasing strong generalization and practical applicability. Our work offers a novel theoretical perspective and practical solution for efficient diffusion model distillation.